\documentclass{article}

\usepackage[final]{neurips_2019}

\usepackage[utf8]{inputenc} 
\usepackage[T1]{fontenc}    
\usepackage{hyperref}       
\usepackage{url}            
\usepackage{booktabs}       
\usepackage{amsfonts}       
\usepackage{nicefrac}       
\usepackage{microtype}      
\usepackage{graphicx}
\usepackage{wrapfig}
\usepackage{algorithm}
\usepackage{algpseudocode}

\title{Shaping representation through communication: Community size effect in artificial learning systems}

\author{
Olivier Tieleman \\ Deepmind \\ {\tt tieleman@google.com} \\\And
Angeliki Lazaridou \\ Deepmind \\ {\tt angeliki@google.com} \\\And
Shibl Mourad \\ Deepmind \\ {\tt shibl@google.com} \\\And
Charles Blundell \\ Deepmind \\ {\tt cblundell@google.com} \\\And
Doina Precup \\ Deepmind \\ {\tt doinap@google.com} \\\And
}
\date{}

\begin{document}
\maketitle
\begin{abstract}
  Motivated by theories of language and communication that explain why communities with large numbers of speakers have, on average, simpler languages with more regularity, we cast the representation learning problem in terms of learning to \textit{communicate}. Our starting point sees the traditional autoencoder setup as a single encoder with a fixed decoder partner that must learn to communicate. Generalizing from there, we introduce \textit{community}-based autoencoders in which multiple encoders and decoders collectively learn representations by being randomly paired up on successive training iterations. We find that increasing community sizes reduce idiosyncrasies in the learned codes, resulting in representations that better encode concept categories and correlate with human feature norms.
\end{abstract}

\section{Introduction}
\label{sec:intro}

Human languages and their properties are greatly affected by the size of their linguistic community~\citep{Reali:etal:2018,Wray:Grace:2007,Trudgill:2011,Lupyan:Dale:2010}. Small linguistic communities of speakers tend to develop more structurally complex languages, while larger communities give rise to simpler languages~\citep{wals}. Moreover, we observe structural simplification as the effective number of speakers grows, as in the example of English language~\citep{McWhorter:2002}. A similar relation between number of speakers and linguistic complexity can also be observed during \textit{linguistic communication}. Speakers, aiming at maximizing communication effectiveness, adapt and shape their conceptualizations to account for the needs of their specific partners, a phenomenon termed in dialogue research as \textit{partner specificity}~\citep{Brennan:Hanna:2009}. As such, speakers and listeners form \textit{conceptual pacts}~\citep{Brennan:Clark:1996}, and in some extreme cases, these pacts are so ad-hoc and idiosyncratic that overhearers cannot follow the discussion~\citep{Schober:Clark:1989}.

In this paper, we try to understand whether the community size effect is unique to humans and natural language, or whether it also emerges in artificial learning systems. 

More specifically, we investigate whether \textit{community-based learners}, i.e.~learners that communicate with a multitude of partners (rather than with a specific one), will shape the representations they communicate to be simpler in nature. In particular, we introduce \textit{community-based autoencoders} (CbAEs), in which there exist multiple encoders and decoders. At every training iteration, one of each is sampled; this pair then performs a traditional autoencoder (AE) training step. Given that the identity of the decoder is not revealed to the encoder during the encoding of the input, the induced representation should be such that all decoders can use it to successfully reconstruct the input. A similar argument holds for the decoder, which at reconstruction time does not have access to the identity of the encoder. We conjecture that this process will reduce the level of idiosyncrasy, resulting in more abstract representations.

We apply CbAEs to two standard computer vision datasets and probe their representations along two axes, testing whether the community size effect results in learners that communicate \textit{abstract} information of the images, such as \textit{concepts} and their \textit{properties}, rather than idiosyncratic and low-level visual information. We find that in contrast to representations induced within a traditional AE framework 1) the CbAE-induced representations encode concept-centric information that can be decoded by a linear classifier and 2) the underlying topology of the CbAE representations of concepts correlates better with human feature norms.

\section{Community-based autoencoders}
\label{sec:autoencoders}

One of the simplest and most widely used ways to do representation learning is to train an autoencoder, i.e., encode the input ${\bf x}$, usually in a  lower-dimensional representation, ${\bf z}=e({\bf x}, {\bf \theta})$ using some parameters ${\bf \theta}$, and then use the ${\bf z}$ representation to decode back the input ${\bf x'}=d({\bf z}, {\bf \phi})$ through another set of parameters ${\bf \phi}$. $\theta$ and $\phi$ are trained by minimizing a reconstruction loss, e.g.,:
    \begin{equation}\label{eq:ae_loss}
        L({\bf x}, {\bf x'}) = ||{\bf x}-{\bf x'}||_2 = ||{\bf x} - d(e({\bf x}, {\bf \theta}), {\bf \phi})||_2 
    \end{equation}
The resulting latent vector ${\bf z}$ is then treated as the induced representation of the input data.

\begin{wrapfigure}{r}{6.5cm}
    \includegraphics[width=\linewidth]{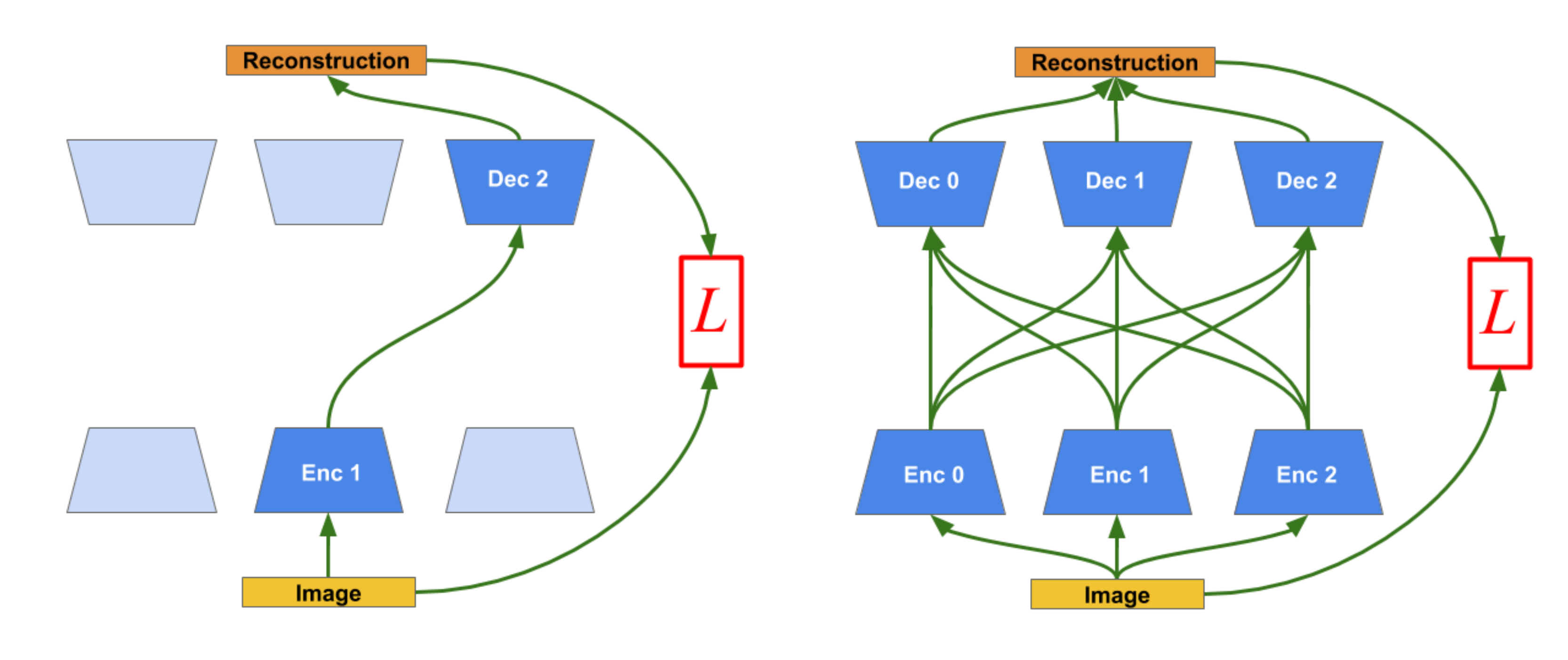}
    \caption{Left: for each iteration, a randomly selected encoder-decoder pair is used. Right: in expectation, all encoders are trained with all decoders, and vice versa.}
    \label{fig:CAE}
\end{wrapfigure}

The CbAE framework (see Figure~\ref{fig:CAE}) is inspired by the hypothesis that the size of a linguistic community has a causal effect on the structural properties of its language. 
Unlike the traditional autoencoder framework, which uses  a single encoder paired with a single decoder, the CbAE set-up involves a collection of $K_{\rm enc}\times K_{\rm dec}$ autoencoders, the outer product of a community of $K_{\rm enc}$ encoders and $K_{\rm dec}$ decoders.\footnote{For simplicity, in our experiments, we use $K_{\rm enc} = K_{\rm dec}$.}
Importantly, while the network architectures can be (and in fact in this work are) identical across a community (all encoders/decoders have the same number and organization of weights) there is no weight-sharing within a community.

\paragraph{Training procedure} At each training step, given a data point ${\bf x}$, we form an autoencoder by randomly sampling an encoder and a decoder from the respective communities. Then, we perform a traditional autoencoding step where we minimize the mean-squared ($L_2$) loss between the input ${\bf x}$ and its decoding (see Eq.~\ref{eq:ae_loss} and Algorithm \ref{alg:cae}). Trivially, the traditional autoencoder training protocol is recovered by setting $K_{\rm enc} = K_{\rm dec} =1$.

\begin{algorithm}
\small
    \caption{Community-based autoencoders}
    \begin{algorithmic}[H]
        \State initialize encoders $\mathcal{E} = \{e_0, ..., e_{K_{\rm enc}}\}$
        \State initialize decoders $\mathcal{D} = \{e_0, ..., e_{K_{\rm dec}}\}$
        \For{each iteration $i$}
            \State sample input data ${\bf x}_i$
            \State sample encoder $e_i$ from $\mathcal{E}$ and decoder $d_i$ from $\mathcal{D}$
            \State ${\bf x'}_i \leftarrow d_i(e_i({\bf x}_i))$
            \State $L_i \leftarrow L({\bf x}'_i, {\bf x}_i)$ \Comment{see Eq.~\ref{eq:ae_loss}}
            \State optimize $e_i$ and $d_i$ with respect to $L_i$
        \EndFor
    \end{algorithmic}
    \label{alg:cae}
\end{algorithm}

There are two main reasons why we think this will have a positive effect on the quality of the representations. First, given that the chosen encoder $e_i$ for iteration $i$ does not have \textit{a priori} information about the identity of the chosen decoder $d_i$, and given that there are a number of decoders all with different weights, the encoder should produce a latent $z_i$ that is decodable by all different decoders. Similarly, given that each decoder $d_i$  receives over its training lifetime latents from a number of different encoders, the decoder should learn to decode representations produced by all encoders. We hypothesize that this training regime will produce latents that are less prone to have idiosyncrasies rooted in the co-adaptation between a particular encoder and decoder.

\paragraph{The curse of co-adaptation}
The goal of our method is to avoid co-adaptation between the encoder and decoder, in a similar way that stable communication partners co-adapt forming ad-hoc communication protocols~\cite{Schober:Clark:1989}. However, due to their flexibility, neural networks are in principle capable of co-adapting to several partner modules at once. As a consequence, the encoders can avoid convergence and still learn to produce latents from which the decoders can successfully reconstruct the input by capitalizing on encoder-specific information. Intuitively, we can think of this as the encoder essentially ``signing'' the latents with their unique ID. We test whether this indeed manifests in the setup by training a linear classifier whose task is to identify the encoder from the latent representation: $p_e(\mathbf{z}) = \exp(\mathbf{w}_e^T \mathbf{z}) / \sum_{e'} \exp(\mathbf{w}_{e'}^T \mathbf{z})$.

\begin{wraptable}{r}{6.7cm}
\begin{tabular}{l|c|c|c|c}

     & \multicolumn{4}{c}{\textbf{Community size}}\\
            & 2 & 4 & 8 & 16 \\
     \hline
     \textit{Chance} & \textit{0.5} & \textit{0.25} & \textit{0.125} & \textit{0.063}\\\hline\hline
     CIFAR-100 & 1.0 & 1.0 & 1.0 & 1.0 \\
     MNIST & 0.84 & 0.37 & 0.19 & 0.09 \\

\end{tabular}
\caption{Encoder identification accuracies.}
\label{tab:encoder_id}
\end{wraptable}

Table~\ref{tab:encoder_id} shows the encoder classification results. The better than random accuracy indicates that pairwise co-adaptation does indeed occur. Increasing the community size can reduce the degree of co-adaptation, since in MNIST we see accuracy drop to levels close to chance for large communities. However, it does not fully overcome the co-adaptation problem, especially in more diverse datasets such as CIFAR-100.

A phenomenon similar to co-adaptation is often encountered in domain-adaption neural frameworks. To alleviate this, adversarial losses or gradient reversal layers~\citep{Ganin:etal:2016} are introduced to penalize representations for retaining domain-specific information. To counteract the pairwise co-adaptation effect, we add to Eq.~\ref{eq:ae_loss} the negative entropy of the encoder classifier (while keeping the latter fixed), to force the encoders to be indistinguishable: $L_{\rm entropy}(\mathbf{z}) = \sum_{e} p_e(\mathbf{z}) \log p_e(\mathbf{z})$.

\paragraph{Training of CbAE}
We use MNIST with community sizes of [1, 2, 4, 8, 16, 32] and CIFAR-100 with community sizes [1, 2, 4, 8] (omitting the larger community sizes due to the size of the networks used). The batch size is fixed at 128 throughout all experiments. We use the Adam optimizer with a learning rate of $10^{-4}$. For MNIST, we use a straightforward 6-layer convolutional neural network of VGG-flavour as encoder, with [64, 64, 128, 128, 128, 128] channels, square kernels of size 3 throughout, and stride 2 for all layers except the first. For CIFAR-100 we use the wide residual network described in \cite{ZagoruykoKomodakis2016} followed by a linear layer with 256 units. The decoders implement the corresponding transpose networks.

\section{Probing the CbAE representations}
\label{sec:representations}

\subsection{Decoding concepts with linear classifiers}
\label{subsec:probe_setup}

We investigate whether increasing the community size results in representations of higher abstraction, which better encode concept-level information about the images, rather than low-level visual information. 
Concretely, after training the CbAE encoders, we train diagnostic classifiers like in e.g.~\cite{Higgins:etal:2017}, one for each encoder.
We fit a linear layer, followed by a softmax, on the latents of each CbAE-trained encoder, predicting the image label. The classifiers are trained with the Adam optimizer with a learning rate of $10^{-3}$ and a minibatch size of 128 throughout all experiments. Note that in this probe task, only the classifiers are trained: the encoders and decoders are frozen.

\paragraph{Results}

\begin{figure*}[th]
    \includegraphics[width=.33\textwidth]{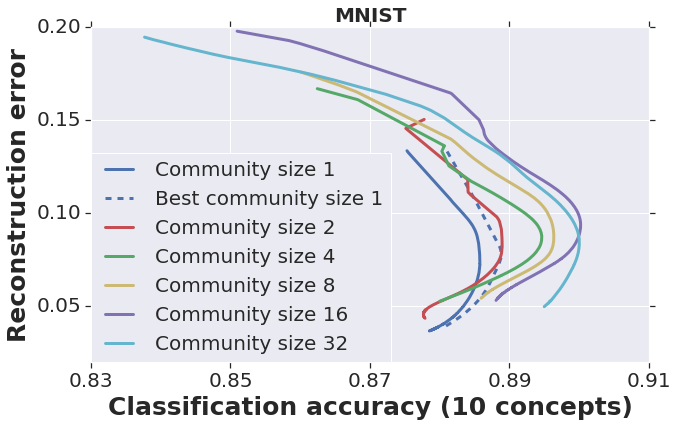}
    \includegraphics[width=.33\textwidth]{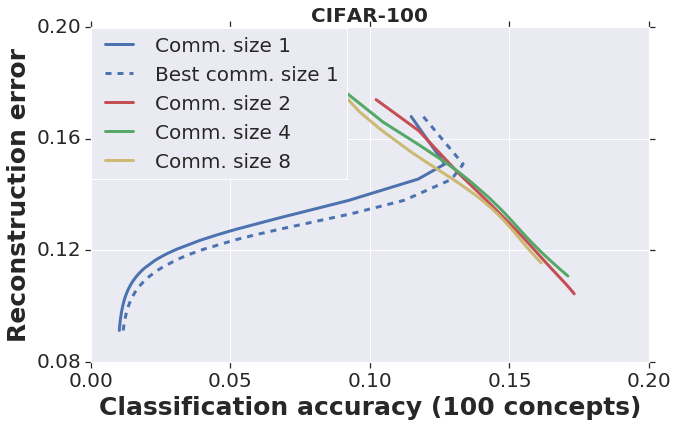}
    \includegraphics[width=.33\textwidth]{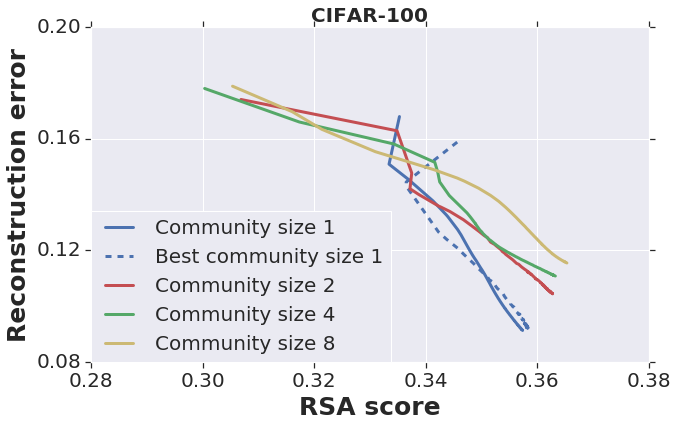}
    \caption{Results as a function of reconstruction error: MNIST concept classification (left), CIFAR-100 concept classification (middle), CIFAR-100 RSA (right). Dashed blue lines indicate the best-performing seed in an ensemble of 32 (MNIST) or 8 (CIFAR-100).
    }
    \label{fig:results_mnist}
\end{figure*}

For each community configuration we report the mean concept accuracy, obtained by averaging the concept accuracies produced by each the linear classifier on top of the CbAE representations. In Figures~\ref{fig:results_mnist} (left) and ~\ref{fig:results_mnist} (middle) we plot this quantity as a function of the reconstruction loss achieved at different points of the CbAE training on MNIST and CIFAR-100 respectively. 
First of all, we observe that having many communication partners results in higher reconstruction error, i.e., accurately communicating detailed low-level pixel information is harder as the community size grows. However, the interesting question is: given the capacity of the latents, are we trading off detailed pixel-level reconstruction performance for other more relevant higher-level properties? 

The classification results do suggest that more higher-level conceptual properties are present in the community-trained latent spaces. In the MNIST experiments, there is a largely monotonic shift towards better concept classification as a function of community size.
On CIFAR-100, we see a more binary effect -- communities of size larger than one clearly have a positive effect, but there does not appear to be further gain from increasing the community size beyond 2-4 members.

However, given that the network capacity grows linearly with the community size, we need to compare against a baseline that has the same capacity (and diversity of initializations) and only lacks the community aspect. Thus, we train an ensemble of 32 independent AEs (each with a different seed) and report both mean (in solid) and best (in dashed) performance in the ensemble (in the case of CIFAR-100 the ensemble size was 8 - like the largest community in that experiment).

The results show that training an encoder within a community of diverse partners can lead to more abstract representations than training a diverse set of independent encoders each with a fixed decoder partner. Thus, it is the community-based aspect of the training that leads to the increased performance. 

While the gains in absolute numbers are small, community-based training is robust across seeds; in the Appendix we show the variability of performance between different CbAE training seeds.

\subsection{Representational similarity analysis}
\label{sec:RSA}

We investigate whether increasing the community size induces abstract representations that share the underlying \textit{structure} of human perceptual data.
As a proxy for the latter, we use the Visual Attributes for Concepts Dataset (VisA) of \citet{Silberer:etal:2013},
containing human-generated per-concept attribute annotations for concrete concepts (e.g., {\ttfamily cat}, {\ttfamily chair}) spanning across different categories
(e.g., {\ttfamily mammals}, {\ttfamily furniture}), annotated with general visual attributes (e.g., {\ttfamily has\_whiskers}, {\ttfamily has\_seat}).

We conduct this experiment only on the CIFAR-100 dataset which contains categories for common nouns, keeping only the 68 of these categories that also occur in ViSA. For measuring the similarities between the VisA and the CbAE-induced representations, we perform Representational Similarity Analysis (RSA) in the two topologies, a method popular in neuroscience~\citep{Kriegeskorte:etal:2008}.
For each community configuration, we sample 5,000 images and encode them with all encoders. For each encoder-specific set of latents, we apply concept-based late fusion (i.e., we average in a single latent all latents belonging to the same concept), arriving to 68 concept  representations. We then compute two sets of pairwise cosine similarities of the 68 concepts in both the ViSA and CbAE spaces, and then  compute the Spearman correlation of these two lists of similarities.

\paragraph{Results} 

For each community configuration we report the mean RSA performance, obtained by averaging the RSA scores produced by the different encoders.
In Figure~\ref{fig:results_mnist} (right) we plot this quantity as a function of the reconstruction loss achieved at different points of the CbAE training.
Again, just like the ensemble control experiment in Section~\label{subsec:probe_setup}, we plot the mean (solid) and maximum (dashed) RSA score in the ensemble of the 8 independent CbAEs of community size 1.

The main result is that the similarity increases with the size of the community. This confirms the hypothesis that CbAE produce on average abstract representations that to some degree reflect the topology of the highly structured and disentangled human feature norm data. 

\section{Discussion}

Motivated  by  theories  of  language  and  communication   that   explain   why   communities with  large  numbers  of  speakers  have simpler  languages  with  more  regularity, in this work we tested whether a similar phenomenon can be observed in artificial learning systems.  Analogous to the structural simplicity found in languages with many speakers, we find that the latent representations induced in this scheme are \textit{more abstract} and \textit{structured}.
We believe these preliminary results  could open avenues to potential synergies between linguistics and representation learning.

\bibliography{emnlp-ijcnlp-2019}
\bibliographystyle{acl_natbib}

\clearpage

\appendix

\section{Supplemental material}

\begin{figure*}[th]
    \includegraphics[width=.33\textwidth]{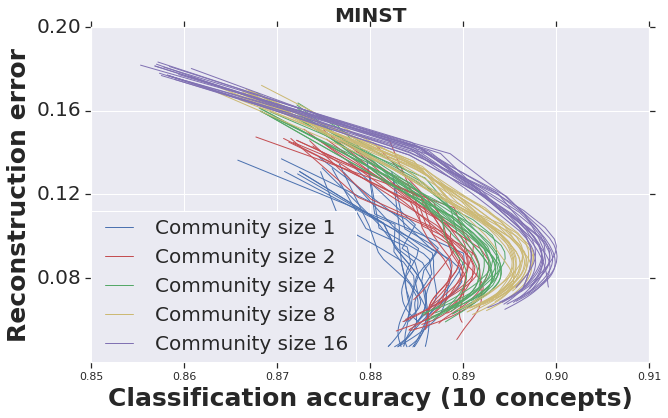}
    \includegraphics[width=.33\textwidth]{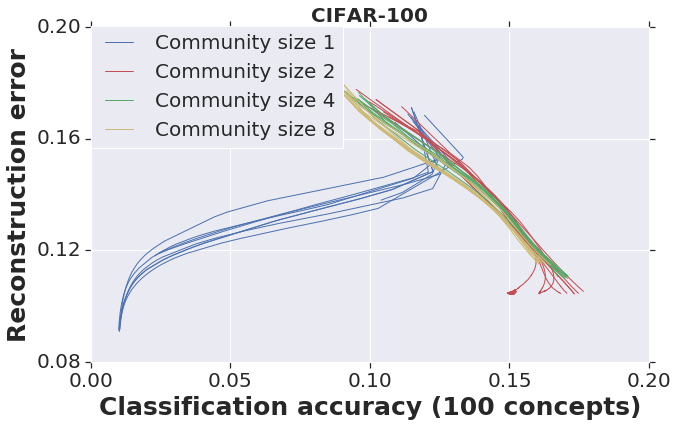}
    \includegraphics[width=.33\textwidth]{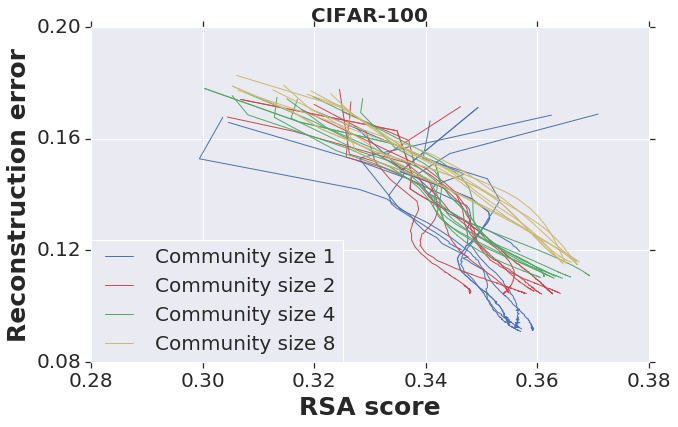}
    \caption{The individual seeds.}
    \label{fig:individual_seeds}
\end{figure*}
    
In figure \ref{fig:individual_seeds} we show the variability between different seeds for the experiments discussed in the main text. Note that in all cases, a larger community leads to less variation between seeds. This effect is a natural consequence of the fact that the individuals within a community need to agree on a shared representation.

\end{document}